\documentclass[runningheads]{llncs}
\usepackage{graphicx}
\usepackage[misc]{ifsym}
\usepackage{algorithm}
\usepackage{algpseudocode}
\usepackage{amsmath,amssymb,amsfonts}
\usepackage{bbding} 
\usepackage{multirow,multicol}
\usepackage{xcolor}
\usepackage{booktabs}
\usepackage[colorlinks,linkcolor=green,citecolor=green]{hyperref}
\usepackage{subfigure}
\usepackage{caption}
\usepackage{graphicx}
\usepackage{float} 
\usepackage{subcaption}
\usepackage{cite}
\usepackage{caption}
\begin{document}
\title{Balancing Complementarity and Consistency via Delayed Activation in Incomplete Multi-view Clustering
}

\titlerunning{CoCo-IMC}
%

\author{Bo Li\inst{1,2} \href{https://orcid.org/0009-0003-7011-4134}{\includegraphics[scale=0.08]{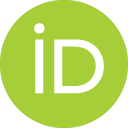}} \and
Zhiwei Xu\inst{2,3} \and
Jing Yun\inst{1{( \textrm{\Letter} )}} \and
Jiatai Wang\inst{2} \href{https://orcid.org/0000-0001-7373-7706}{\includegraphics[scale=0.08]{pic/orcid.png}}
}
\authorrunning{B. Li et al.}

\institute{
College of Data Science and Application, Inner Mongolia University of Technology, Huhhot, China\\
\email{yunjing\_zoe@163.com} \\ \and
Haihe Lab of ITAI, Tianjin, China \and
Institute of Computing Technology, Chinese Academy of Sciences, Beijing, China\\}

%
%
\maketitle              
\begin{abstract}
This paper study one challenging issue in incomplete multi-view clustering, where valuable complementary information from other views is always ignored. To be specific, we propose a framework that effectively balances \underline{Co}mplementarity and \underline{Co}nsistency information in \underline{I}ncomplete \underline{M}ulti-view \underline{C}lustering (CoCo-IMC).
Specifically, we design a dual network of delayed activation, which achieves a balance of complementarity and consistency among different views. The delayed activation could enriches the complementarity information that was ignored during consistency learning.
Then, we recover the incomplete information and enhance the consistency learning by minimizing the conditional entropy and maximizing the mutual information across different views. This could be the first theoretical attempt to incorporate delayed activation into incomplete data recovery and the balance of complementarity and consistency.
We have proved the effectiveness of CoCo-IMC in extensive comparative experiments with 12 state-of-the-art baselines on four publicly available datasets.

\end{abstract}
\section{INTRODUCTION}
\label{sec1}
\begin{figure}
\centering
\begin{minipage}{0.19\linewidth}
    \centerline{\includegraphics[width=1.1\textwidth]{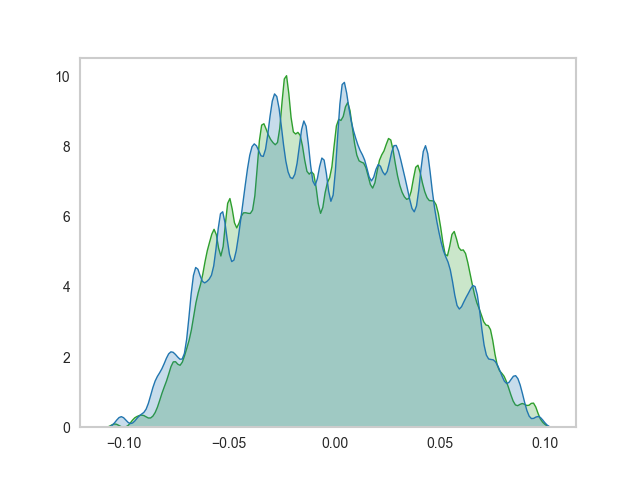}}
    \centerline{\includegraphics[width=1.1\textwidth]{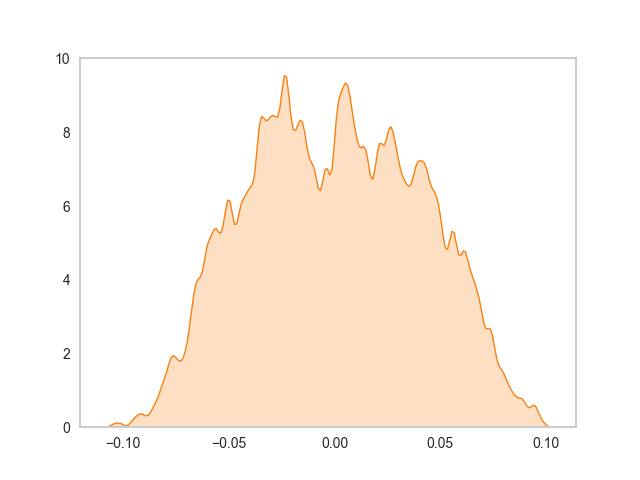}}
    \centerline{(a)}
\end{minipage}
\begin{minipage}{0.19\linewidth}
    \centerline{\includegraphics[width=1.1\textwidth]{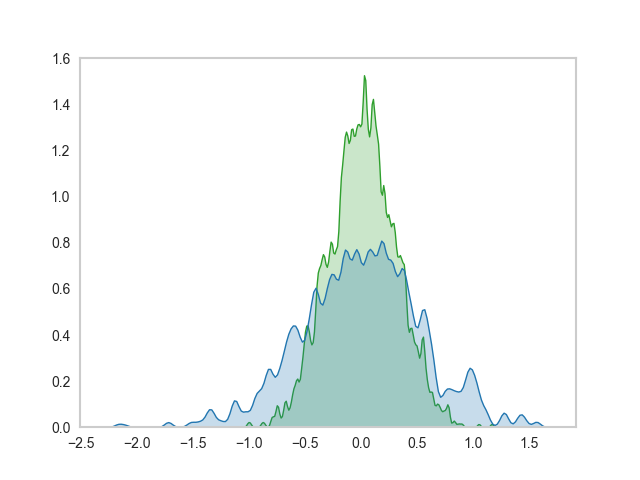}}
    \centerline{\includegraphics[width=1.1\textwidth]{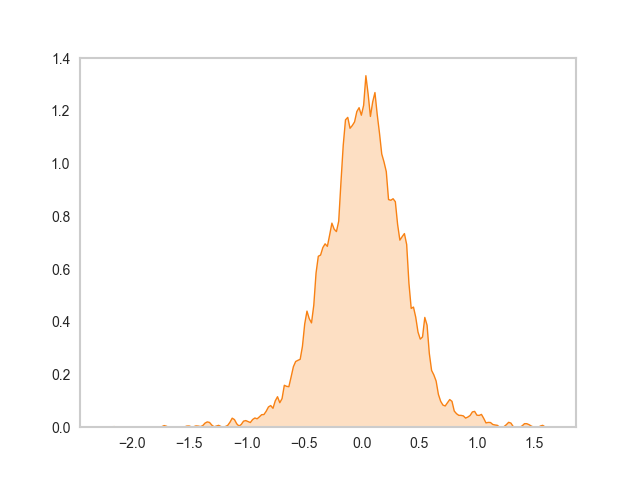}}
    \centerline{(b)}
\end{minipage}
\begin{minipage}{0.19\linewidth}
    \centerline{\includegraphics[width=1.1\textwidth]{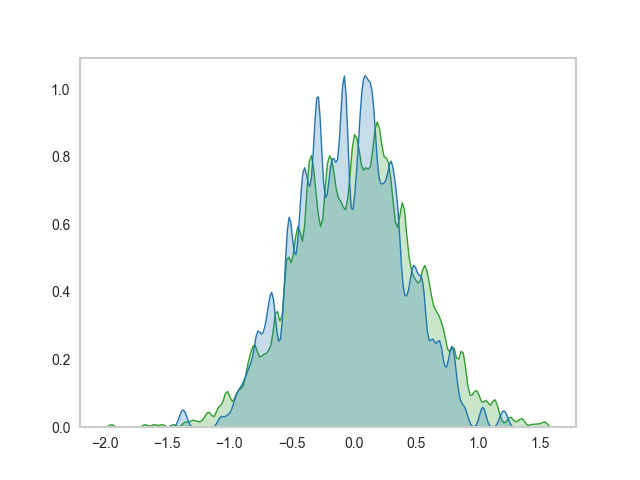}}
    \centerline{\includegraphics[width=1.1\textwidth]{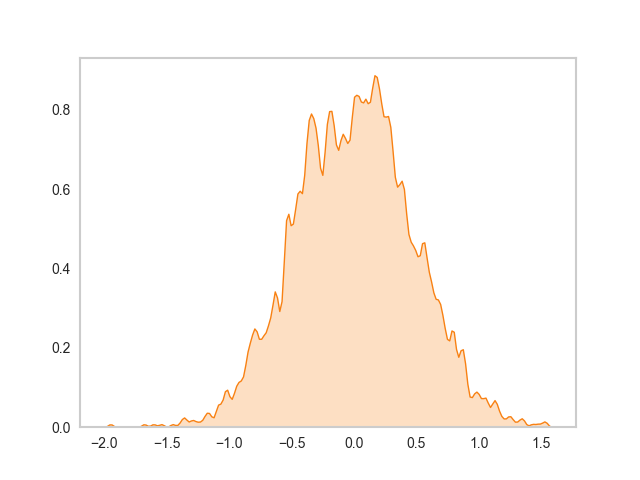}}
    \centerline{(c)}
\end{minipage}
\begin{minipage}{0.19\linewidth}
    \centerline{\includegraphics[width=1.1\textwidth]{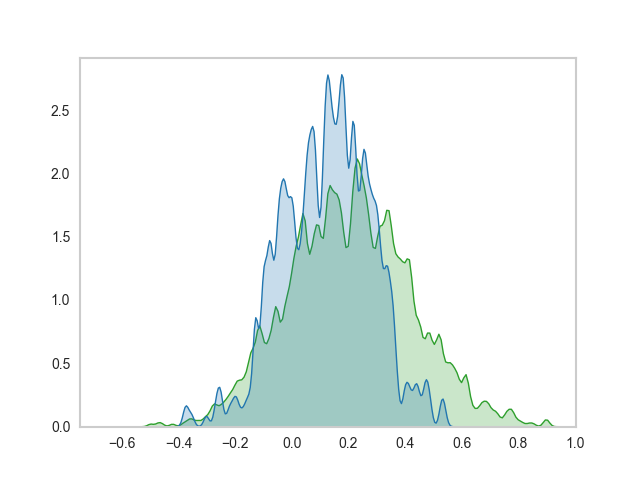}}
    \centerline{\includegraphics[width=1.1\textwidth]{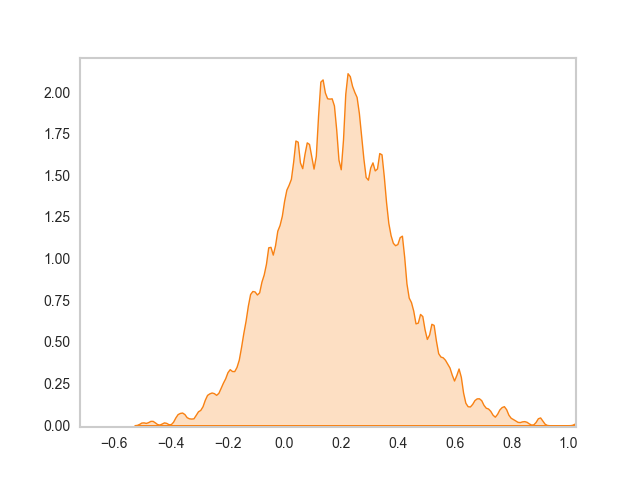}}
    \centerline{(d)}
\end{minipage}
\begin{minipage}{0.19\linewidth}
    \centerline{\includegraphics[width=1.1\textwidth]{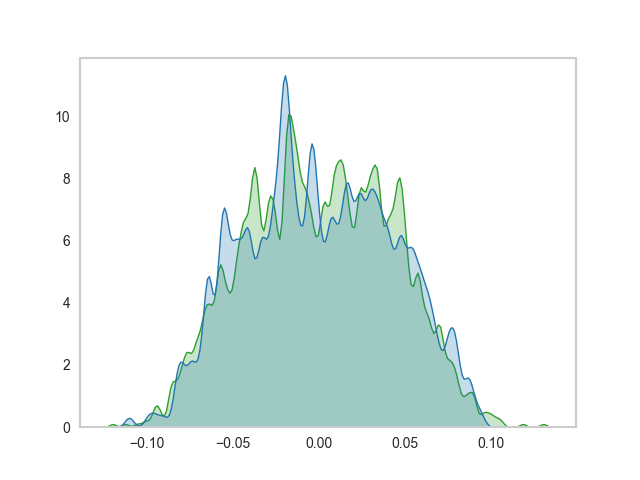}}
    \centerline{\includegraphics[width=1.1\textwidth]{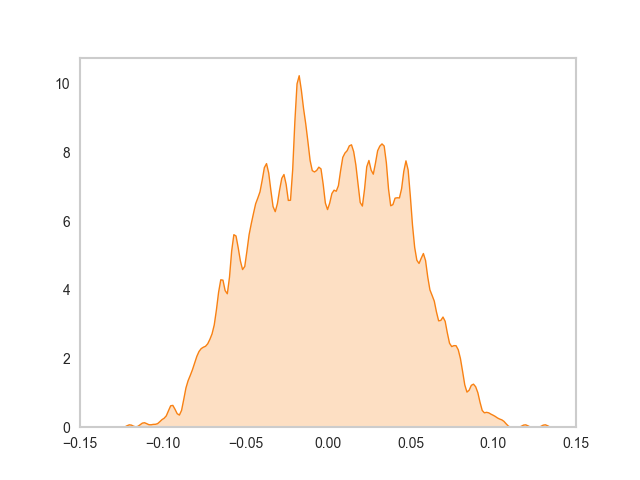}}
    \centerline{(e)\textbf{Ours}}
\end{minipage}
\caption{\small{Illustration of  the motivation and the experimental results from multi-view data distribution on Caltech101-20. (a) Raw distribution. (b) Missing distribution. (c) Missing distribution with consistency. (d) Missing distribution with consistency and complementary. (e) Missing distribution with balancing consistency and complementary (\textbf{Ours}). In the figure, the green and blue curves denote the data distribution of two view, respectively. The orange curves denote the true distribution. We could observe that achieving a balance between complementarity and consistency in incomplete multi-view can closer to the true semantic, which is the most desirable result of multi-view clustering.} }\label{motivation}
\vspace{-0.6cm}
\end{figure}

Multi-view data is collected from diverse sensors or obtained from various feature extractors\cite{yan2023gcfagg}. For example, a patient's lesions are often represented by magnetic resonance imaging(MRI) and positron emission tomography(PET). However, data heterogeneity can increase the difficulty of mining common semantics. Multi-view clustering(MVC) aim at exploiting correlation across different views for data analysis (See Fig. \ref{motivation} (a)). Existing multi-view clustering methods rely on consistency among different views to learn a common representation. Missing some multi-view data is often unavoidable during data collection and transmission, thus leading to incomplete multi-view clustering(IMC) (See Fig. \ref{motivation} (b)). 

Recently, some efforts have recovered incomplete data by utilizing the information of the existing views, which alleviates the impacts of incomplete information and ensures consistency learning. Yan et al.\cite{yan2024multi} learn more discriminative consistency representations by information bottleneck theory to reduce the adverse effects of complementarity. In addition, some works enhance consistency learning by introducing contrastive learning, which achieves excellent performance. Yang et al.\cite{SURE} impute the missing sample by its peers in the same view by robust contrastive learning. Such methods pay attention to the consistency of all views in the process of consistency learning. However, complementary information as helpful extra information is ignored, which may be very important.

To utilize complementary information, typical works actively enhance or indicately preserve complementarity. Zhang et al.\cite{69} forced the learned similarity matrices to have complementarity by introducing the Hilbert independence criterion. However, utilizing complementary information directly will result in redundancy of information and worse clustering results (See Fig. \ref{motivation} (d)). Therefore, it is essential to achieve a balance of complementarity and consistency in incomplete multi-view clustering.

Based on the above fact, we propose a novel incomplete multi-view clustering method, CoCo-IMC, that implements a balance of complementarity and consistency. As shown in Fig. \ref{fig1}. We first design a dual network of delayed activation, which forces the model to concentrate on the features of the training data preceding the current time point. This learning process gives the model the opportunity to understand the data from multiple perspectives, rather than relying on the most notable features. Thus, the model is able to capture more subtle information and learn more complementary representations (See Fig. \ref{motivation} (e)). After that, CoCo-IMC recovers the incomplete information and enhances consistency learning by minimizing the conditional entropy and maximizing the mutual information among different views. The delayed activation prevents the model from adapting to the training data too quickly and falling into a local optimum, which directly enhances the consistency learning. Moreover, the retained complementarity information from the delayed activation provides more comprehensive data support for clustering, which indirectly enhances consistency learning. The main contributions of this paper are:

\begin{itemize}
    \item[$\bullet$] We find that complementarity information contributes to incomplete multi-view clustering, which is usually ignored.
    
    \item[$\bullet$] Based on this finding, we propose a novel incomplete multi-view clustering method, CoCo-IMC, that achieves the balance between complementarity and consistency of multi-view by delayed activation in a unify framework.
    
    \item[$\bullet$] Extensive experiments demonstrate that CoCo-IMC achieves excellent clustering performance on four publicly available datasets compared to 12 state-of-the-art IMC methods. 
\end{itemize}
\section{RELATED WORKS}
\label{sec2}
\subsection{Multi‐view Clustering}
\label{sec2.1}
Recently, research on multi-view clustering has been done by implicitly or explicitly exploiting the common semantics between data from different views\cite{90,91,92,wang2023hierarchical}. Existing MVC methods can be broadly divided into five categories, i.e., matrix factorization-based MVC, spectral clustering-based MVC, kernel learning-based MVC, graph learning-based MVC and deep learning-based MVC. Matrix factorization-based MVC methods usually utilize low rank to project the data into the embedding space to find common latent factors to achieve multi-view clustering. DAIMC\cite{DAIMC}uses $\mathcal{L}_{2,1}$-norm to establish a consensus basic matrix. IMG \cite{IMG} utilizes the $\mathcal{L}_{F}$-norm to alleviate the effects of missing data. Likewise, kernel learning-based MVC usually uses a multi-kernel learning to map each view into a kernel matrix \cite{87}. EERIMVC \cite{EERIMVC} achieves multi-view clustering by iterative optimization multi-kernel methods. However, the above methods are normally unable to explore nonlinear data structures. Graph learning-based methods remove this limitation by exploring the fused graph. PIC \cite{PIC} use to construct consistent Laplacian plots from incomplete views to handle nonlinear data.
Although the above methods utilize embedding functions to capture the nonlinear nature of multi-view, there are still limitations in terms of their representation capabilities and computational complexity.
Deep learning-based MVC has attracted researchers due to their powerful feature extraction capabilities. Wang et al.\cite{CIMIC-GAN} uses generative adversarial network to solve incomplete multi-view problem. For further MVC methods, please refer to the relevant surveys\cite{54}.

The differences between this paper and existing works are given blow. We aim to balance the complementarity and consistency of multi-view. Moreover, our method is based on deep learning, and we adopt a novel dual-network structure to explore the feature information of views. Note that the delayed activation of the dual network prevents the model from adapting to the training data too quickly, which directly enhances consistency learning. And the retained complementarity information indirectly enhances consistency learning.

\subsection{Contrastive Learning}
\label{sec2.2}
As one of the essential unsupervised learning paradigms, contrastive learning is extremely valuable in capturing correlations between samples\cite{33}. Its primary objective is to maximize the similarity between similar samples while minimizing the similarity between different samples in the latent space, which enhances the model's ability to recognize samples and facilitates more accurate classification\cite{104}. Existing methods can be broadly divided into two categories. The first category involves minimizing InfoNCE losses. Typical CPC\cite{29} and CMC\cite{34} explicitly defines positive and negative sample pairs to contrast their differences. MoCo\cite{32} and SimCLR\cite{23} continue this idea. In order to maintain a sufficient number of negative samples, they respectively adopt a momentum update mechanism and introduce up to ten forms of data augmentation. The second category is prediction-based methods. Compared to being limited by the definition of negative sample pairs, BYOL\cite{25}, SimSiam\cite{35} and DINO\cite{36} completely abandon negative sample pairs, successfully convert the contrastion task into a prediction task, and achieve excellent results. SwAV\cite{30} promotes consistency of the same sample between different views.

Existing contrastive learning MVC methods aim to construct various data augmentations to emphasize the importance of consistency learning in MVC. Instead, CoCo-IMC balanced the complementarity and consistency information in multi-view to effectively enhance the true semantic of multi-view.
\section{METHODS}
\label{sec3}
\begin{figure}[ht!]
\centering
\includegraphics[width=0.8\textwidth]{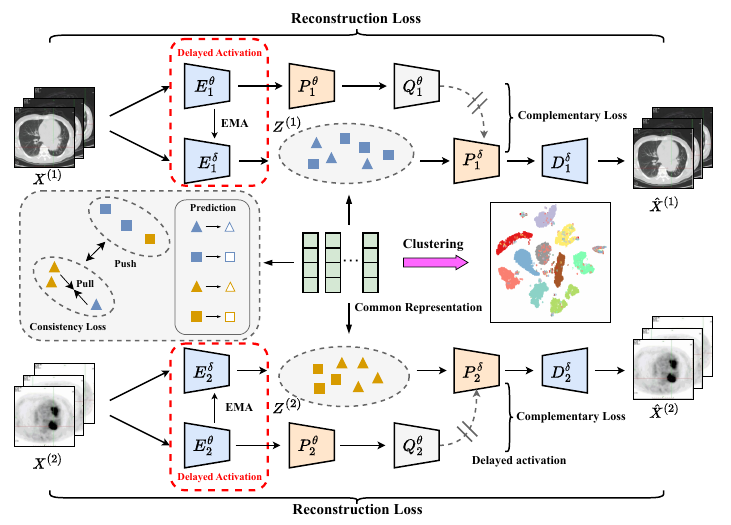}
\caption{\small{Overview of CoCo-IMC. Bi-view data is used as a showcase in this figure. CoCo-IMC consists of three joint learning objectives, i.e., complementarity learning, consistency learning and reconstruction. Specifically, the complementary learning objective is to capture and balance the information of complementary and consistency. Consistency learning is to maximize the mutual information among different views. The goal of reconstruction is to project all views into a specific space.
}} \label{fig1}
\vspace{-0.4cm}
\end{figure}

\vspace{-0.2cm}
In this section, we propose a novel deep incomplete multi-view clustering method, CoCo-IMC, to balance the complementarity and consistency of multi-view. As illustrated in Fig. \ref{fig1}, CoCo-IMC consists of three jointly learning objectives, namely complementarity learning, consistency learning and view reconstruction. For clarity, we will first introduce the proposed loss function and then elaborate on each objective.
\vspace{-0.2cm}

\subsection{Notations}
\label{sec3.1}
Given a multi-view dataset $X^{(v)}=\left \{ x_1^{(v)},x_2^{(v)},\dots, x_n^{(v)} \right\} \in \mathbb{R}^{m_v \times n}$ includes $n$ samples across $v$ views, where $m_v$ is the feature dimension and $n$ represents the number of instances. The missing rate of data is defined as $\eta = \frac{n-m}{n}(m \le n)$, where $m$ is the number of complete samples. Incomplete mult-view clustering attempts to solve the problem of missing views in instances by dividing $n$ instance samples into $k$ clusters.

With the above definitions, we propose the following objective function:
\begin{equation}
\label{eq1}
\mathcal L= \beta \mathcal L_{cml} +\lambda \mathcal L_{cnl} + \alpha \mathcal L_{rec} 
\end{equation}
where $\mathcal L_{cml}$, $\mathcal L_{cnl}$ and $\mathcal L_{rec}$ are complementarity leanring loss, consistency learning loss and view reconstruction loss, respectively. The parameters $\beta$, $\lambda$ and $\alpha$ are the balanced factors on $\mathcal L_{cml}$, $\mathcal L_{cnl}$ and $\mathcal L_{rec}$, respectively. In our experiments, we set these three parameters to 0.001, 1 and 0.01, respectively. We will prove this in the following parametric analysis experiment.

\subsection{Complementary Module}
\label{sec3.2}
\begin{figure}
    \centering
    \begin{minipage}[t]{0.49\linewidth}
        \centering
        \includegraphics[width=1\linewidth]{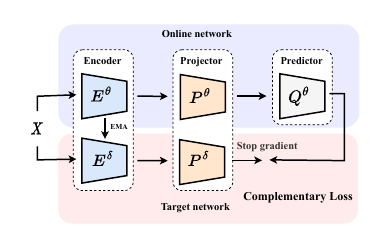}
        \caption{\small{Complementary module.}}
        \label{fig3}
    \end{minipage}
    \begin{minipage}[t]{0.49\linewidth}
        \centering
        \includegraphics[width=1\linewidth]{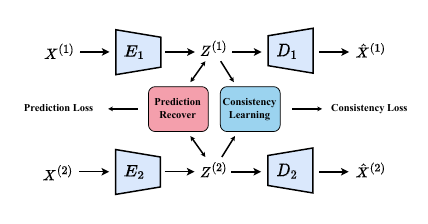}
        \caption{\small{Consistency module.}}
        \label{fig4}
    \end{minipage}
\end{figure}

As shown in Fig. \ref{fig3}, in this module, our dual network firstly achieves feature mining by contrastive learning with no negative samples. This process that maximises feature diversity. Specifically, the dual network consists of an online network and a target network. The online network has an encoder $E^{\theta}$, a projector $P^{\theta}$, and another predictor $Q^{\theta}$, with the weight parameter $\theta$, and the target network has an encoder $E^{\delta}$ and a projector $P^{\delta}$, and the weight parameter is $\delta$. Compared with traditional contrastive learning, we directly use the softmax function in the predictor $Q$ for explicit prediction. This enables the online network to learn more about the differences between different multi-view distributions, while the target network can maintain this difference to a great extent. The outputs of the online network and the target network are consistent by using a mean squared error loss function. The process can be expressed as:
\begin{equation}
\label{eq2}
\mathcal L_{Z^{(1)} \to X^{(1)}} =  \left \|  Z^{(1)}- P(E(X^{(1)})) \right \|_2^2 = 2-2 \cdot \frac{\left \langle  Z^{(1)},P(E(X^{(1)}))\right \rangle}{\left \| Z^{(1)} \right \|_2 \cdot \left \| P(E(X^{(1)})) \right \|_2}
\end{equation}
where $Z^{(1)}=Q(P(f(X^{(1)})))$ denotes the output of $X^{(1)}$ through the online network. After that, we then feed $X^{(1)}$ into the target network to obtain the symmetric loss function $\mathcal L_{X^{(1)} \to Z^{(1)}}$ respectively. Then the complementary loss as follows:
\begin{equation}
\label{eq3}
\mathcal L_{cml}=\mathcal L_{Z^{(1)}\to X^{(1)}}+\mathcal L_{X^{(1)}\to Z^{(1)}}
\end{equation}

Then, we achieve a balance of complementarity and consistency in multi-view clustering through delayed activation of the dual network. Specifically, in the training process, the encoder $E^{\theta}$ is updated with the gradient, and the encoder $E^{\delta}$ is updated in the form of a moving average. Encoder $E^{\delta}$ provides the regression target while training encoder $E^{\delta}$ and its parameters are an exponential moving average of encoder  $E^{\theta}$. Therefore, the model is forced to focus on features of the training data at different points in time during this process, which makes the model understand the data from multiple perspectives rather than rely on the most notable features and learn more complementary information. Finally, the delayed activation achieves a balance of complementarity and consistency in the learning process. The target momentum $m\in \left [ 0,1\right ] $ is updated as follows:
\begin{equation}
\label{eq4}
\delta\gets m \delta  + (1-m)\theta
\end{equation}
Note that if we choose a larger momentum during training, the parameters of the target network are updated more slowly, ensuring the target network remains similar to the online network. Thus, the model retains more complementary information.
\leavevmode\newline\vspace{-1cm}
\leavevmode\newline\vspace{-1cm}

\subsection{Consistency Module}
\label{sec3.3}
Consistency learning across different views and incomplete view recovery are mutually reinforcing processes\cite{COMPLETER}. As shown in Fig. \ref{fig4}, in the $L_{rec}$ parameterized latent space, we explicitly predict view-specific representations by minimizing the conditional entropy $Z_n^{(1)}$ and $Z_n^{(2)}$, then maxizing mutual information across different views. We define the consistency learning loss as:
\begin{equation}
\label{eq5}
\mathcal L_{cnl}= \mathcal L_{pre}+ \mathcal L_{ccl}
\end{equation}

Specifically, we predict the missing representation $Z^{(j)}$ by the existing representation $Z^{(i)}$. Taking the double view as an example, the recovery loss has the following form:
\begin{equation}
\label{eq6}
\mathcal L_{pre} = \left \|  P^{(1)}(Z^{(1)})-Z^{(2)}\right \| _2^2 + \left \|  P^{(2)}(Z^{(2)})-Z^{(1)}\right \| _2^2
\end{equation}

Maximizing the consistency between $Z^{(1)}$ and $Z^{(2)}$ can be expressed as:
\begin{equation}
\label{eq7}
\mathcal L_{ccl} = - \sum_{n=1}^N (I(Z_n^{(1)},Z_n^{(2)})+\eta(H(Z_n^{(1)})+H(Z_n^{(2)})))
\end{equation}
where $\eta$ denotes used to balance the regularization hyperparameters. $H$ denotes the information entropy, and the larger $H$ is, the more information it contains. $I$ denotes mutual information, $Z_n^{(1)}$ and $Z_n^{(2)}$ denote two discrete cluster assignment variables, the output of the softmax function in the decoder is used as the cross-cluster probability distribution of $Z_n^{*}$ \cite{107}, and the joint probability distribution of $Z_n^{(1)}$ and $Z_n^{(2)}$ is obtained. The formula is as:
\begin{equation}
\label{eq8}
\begin{split}
I(Z_n^{(1)},Z_n^{(2)}) & =-\sum_{Z_n^{(1)}}^{N} \sum_{Z_n^{(2)}}^{N} P_{Z_n^{(1)},Z_n^{(2)}} \frac{log(P_{Z_n^{(1)}}P_{Z_n^{(2)}})}{log(P_{Z_n^{(1)}}^{\eta +1}P_{Z_n^{(2)}}^{ \eta +1})}
\end{split}
\end{equation}
where $\eta$ is the regularization factor.

\begin{algorithm}
\caption{CoCo-IMC}
\label{alg:cap}
\begin{algorithmic}
\Require 
\State Multi-view dataset $X_n^{(v)}=\left \{ X_n^{(1)},X_n^{(2)} \right \}$; Learning rate $\eta$; Pre-training epochs $n_1$; Training epochs $n$; Hyper-parameter $\alpha$,$\beta$,$\lambda$; Momentum $m$; Cluster number $K$
\Ensure
\State Pretrain autoencoders $\theta$ and $\delta$ with the corresponding multi-view
\State Train with corresponding incomplete multi-view

\While{$epoch \le n$}
    \State Inintialize autoencoder parameters $\theta$ and $\delta$
    \State Optimize $\theta$ and $\delta$ by minimizing Eq. (\ref{eq3}), Eq. (\ref{eq7}) and Eq. (\ref{eq9})    
\If{$epoch \le n_1$}
    \State Compute prediction loss by Eq. (\ref{eq6})
\EndIf
\EndWhile
\State \textbf{Output:}
\State Transform all data into latent representation $Z_n$; Clustering on the common representation $Z_n$; Obtain the clustering results $R$
\State \textbf{Return $R$}
\end{algorithmic}
\end{algorithm}
\leavevmode\newline\vspace{-1.2cm}

\subsection{Reconstruction Module}
\label{sec3.4}
For each view, we pass it through an autoencoder to learn the latent representation by minimizing:
\begin{equation}
\label{eq9}
\mathcal{L}_{rec} = \sum_{v=1}^{2}\sum_{n=1}^{N} || X_{n}^{(v)} - D^{(v)}(E^{(v)}(X^{(v)}_{n}))||_2^2
\end{equation}
where $X_n^{(v)}$ denotes the $n$-th sample of view $X^{(v)}$. $E^{(v)}$ and $D^{(v)}$ denote the encoder and decoder of the $v$-th view, respectively. Thus, the $n$-th sample in the $v$-th view is represented as:
\begin{equation}
\label{eq10}
Z_{n}^{(v)}=E^{(v)}(X_n^{(v)})
\end{equation}
where $Z^{(v)}$ denotes the latent representation of view $X^{(v)}$, $E^{(v)}$ denotes the encoder of the $v$-th view, and $v \in \left \{1, 2 \right \}$.

It should be pointed out that the autoencoder structure is helpful to avoid the trivial solution.
\section{EXPERIMENTS}
\label{sec4}
In this section, we evaluate the proposed CoCo-IMC method on four widely used multi-view datasets with comparisons of 12 state-of-the-art clustering methods.

\subsection{Experimental Settings}
\label{sec4.1}
Four widely used datasets are used in our experiments including: \textbf{Caltech101-20} consists of 2,386 images of 20 subjects with the views of Histogram of Oriented Gradient (HOG) and Global Feature Information (GIST) features, with 1,984 and 512 as the feature dimensions, respectively. \textbf{Scene-15} consists of 4,485 images distributed over 15 scene categories with the view of GIST and Pyramid Histogram of Oriented Gradient (PHOG) features, 59D and 20D feature vectors, respectively. \textbf{LandUse-21} consists of 2,100 satellite images from 21 categories with PHOG and Local Binary Pattern (LBP) features, 59D and 40D feature vectors, respectively. \textbf{Noisy MNIST} consists of 70k instances of 10 categories. We randomly select 15k original instances as view 1 and 15k Gaussian noise instances as view 2.

To evaluate the performance of CoCo-IMC, three widely used clustering metrics including Accuracy (ACC), Normalized Mutual Information (NMI) and Adjusted Rand Index (ARI)\cite{DAIMC}. A higher value of these metrics indicates better clustering performance.

The experiments are executed using the following hardware configuration: Inter Core i7-12700 CPU, NVIDIA GeForce RTX 3060 GPU and 16GB RAM. Additionally, the PyTorch platform is employed for all experiments. In the case of CoCo-IMC, we utilize the Adam optimizer to minimize the total loss.

\subsection{Comparisons with State of The Arts}
\label{sec4.2}
We compare CoCo-IMC with 12 multi-view methods including $AE^2$-Nets\cite{AE2NETS}, IMG\cite{IMG}, UEAF\cite{UEAF}, DAIMC\cite{DAIMC}, EERIMVC\cite{EERIMVC}, DCCAE\cite{DCCAE}, PVC\cite{PVC}, BMVC\cite{BMVC}, DCCA\cite{DCCA}, PIC\cite{PIC}, COMPLETER\cite{COMPLETER} and CIMIC-GAN\cite{CIMIC-GAN}.
The $AE^2$-Nets, DCCAE, BMVC and DCCA could only handle the complete multi-view data. Thus, we filled the missing data with the mean value of the same view. For all methods, we use the recommended network structure and parameters for a fair comparison.

\vspace{-0.8cm}
\begin{table}
\caption{The clustering performance comparison of two-view datasets in incomplete and complete MVC settings. ‘-’ indicates unavailable results due to out of memory. The ${1^{\mathrm{st}}}$ best results are indicated in \textbf{bold} \color{black}and the ${2^{\text {nd }}}$ best results are indicated in \underline{underlined}. }\label{table4.2}
\renewcommand\arraystretch{1}
\centering
\footnotesize
\setlength{\tabcolsep}{0.8mm}
\scalebox{0.68}{
\begin{tabular}{llcccccccccccc}
\toprule
 & Method\textbackslash{}Datasets & \multicolumn{3}{c}{Caltech101-20} & \multicolumn{3}{c}{Scene-15} & \multicolumn{3}{c}{LandUse-21} & \multicolumn{3}{c}{Noisy MNIST} \\
\multirow{-2}{*}{DataType} & Evaluation metrics & ACC & NMI & ARI & ACC & NMI & ARI & ACC & NMI & ARI & ACC & NMI & ARI \\ \hline
 & AE$^2$Nets\cite{AE2NETS}(2019) & 33.61 & 49.20 & 24.99 & 27.88 & 31.35 & 13.93 & 19.22 & 23.03 & 5.75 & 38.67 & 33.79 & 19.99 \\
 & IMG\cite{IMG}(2016) & 42.29 & 58.26 & 33.69 & 23.96 & 25.70 & 9.21 & 15.52 & 22.54 & 3.73 & - & - & - \\
 & UEAF\cite{UEAF}(2019) & 47.35 & 56.71 & 37.08 & 28.20 & 27.01 & 8.70 & 16.38 & 18.42 & 3.80 & 34.56 & 33.13 & 24.04 \\
 & DAIMC\cite{DAIMC}(2019) & 44.63 & 59.53 & 32.70 & 23.60 & 21.88 & 9.44 & 19.30 & 19.45 & 5.80 & 34.44 & 27.15 & 16.42 \\
 & EERIMVC\cite{EERIMVC}(2020) & 40.66 & 51.38 & 27.91 & 33.10 & 32.11 & 15.91 & 22.14 & 25.18 & 9.10 & 54.97 & 44.91 & 35.94 \\
 & DCCAE\cite{DCCAE}(2015) & 40.01 & 52.88 & 30.00 & 31.75 & 34.42 & 15.80 & 14.94 & 20.94 & 3.67 & 61.79 & 59.49 & 33.49 \\
 & PVC\cite{PVC}(2014) & 41.42 & 56.53 & 31.00 & 25.61 & 25.31 & 11.25 & 21.33 & 23.14 & 8.10 & 35.97 & 27.74 & 16.99 \\
 & BMVC\cite{BMVC}(2018) & 32.13 & 40.58 & 12.20 & 30.91 & 30.23 & 10.93 & 18.76 & 18.73 & 3.70 & 24.36 & 15.11 & 6.50 \\
 & DCCA\cite{DCCA}(2013) & 38.59 & 52.51 & 29.81 & 31.83 & 33.19 & 14.93 & 14.08 & 20.02 & 3.38 & 61.82 & 60.55 & 37.71 \\
 & PIC\cite{PIC}(2019) & 57.53 & 64.32 & 45.22 & 38.70 & 37.98 & 21.16 & 23.60 & 26.52 & 9.45 & - & - & - \\
 & COMPLETER\cite{COMPLETER}(2021) & 68.44 & 67.39 & 75.44 & \underline{39.50} & 42.35 & 23.51 & 22.16 & 27.00 & 10.39 & 80.01 & 75.23 & 70.66 \\
 & CIMIC-GAN\cite{CIMIC-GAN}(2022) & \underline{69.48} & \underline{68.25} & \underline{75.12} & 39.09 & \textbf{46.12} & \underline{23.55} &\underline{23.76} & \underline{28.03} & \underline{11.10} & \underline{81.97} & \underline{77.22} & \underline{72.56} \\
\multirow{-13}{*}{50\%Missing} & \textbf{CoCo-IMC(ours)} & \textbf{76.78}& \textbf{70.51} & \textbf{87.01} & \textbf{43.11} & \underline{43.45} & \textbf{25.66} & \textbf{25.68} & \textbf{31.25} & \textbf{12.78} & \textbf{83.26} & \textbf{79.24} & \textbf{72.81} \\ \hline
 & AE$^2$Nets\cite{AE2NETS}(2019) & 49.10 & 65.38 & 35.66 & 36.10 & 40.39 & 22.08 & 24.79 & 30.36 & 10.35 & 56.98 & 46.83 & 36.98 \\
 & IMG\cite{IMG}(2016) & 44.51 & 61.35 & 35.74 & 24.20 & 25.64 & 9.57 & 16.40 & 27.11 & 5.10 & - & - & - \\
 & UEAF\cite{UEAF}(2019) & 47.40 & 57.90 & 38.98 & 34.37 & 36.69 & 18.52 & 23.00 & 27.05 & 8.79 & 67.33 & 65.37 & 55.81 \\
 & DAIMC\cite{DAIMC}(2019) & 45.48 & 61.79 & 32.40 & 32.09 & 33.55 & 17.42 & 24.35 & 29.35 & 10.26 & 39.18 & 35.69 & 23.65 \\
 & EERIMVC\cite{EERIMVC}(2020) & 43.28 & 55.04 & 30.42 & 39.60 & 38.99 & 22.06 & 24.92 & 29.57 & 12.24 & 65.47 & 57.69 & 49.54 \\
 & DCCAE\cite{DCCAE}(2015) & 44.05 & 59.12 & 34.56 & 36.44 & 39.78 & 21.47 & 15.62 & 24.41 & 4.42 & 81.60 & 84.69 & 70.87 \\
 & PVC\cite{PVC}(2014) & 44.91 & 62.13 & 35.77 & 30.83 & 31.05 & 14.98 & 25.22 & 30.45 & 11.72 & 41.94 & 33.90 & 22.93 \\
 & BMVC\cite{BMVC}(2018) & 42.55 & 63.63 & 32.33 & 40.50 & 41.20 & 24.11 & 25.34 & 28.56 & 11.39 & 81.27 & 76.12 & 71.55 \\
 & DCCA\cite{DCCA}(2013) & 41.89 & 59.14 & 33.39 & 36.18 & 38.92 & 20.87 & 15.51 & 23.15 & 4.43 & 85.53 &  \textbf{89.44} & 81.87 \\
 & PIC\cite{PIC}(2019) & 62.27 & 67.93 & 51.56 & 38.72 & 40.46 & 22.12 & 24.86 & 29.74 & 10.48 & - & - & - \\
 & COMPLETER\cite{COMPLETER}(2021) & 70.18 & 68.06 &  \underline{77.88} & 41.07 & 44.68 & 24.78 & 25.63 &  \underline{31.73} &  \underline{13.05} & 89.08 & \underline{88.86} & 85.47 \\
 & CIMIC-GAN\cite{CIMIC-GAN}(2022) & \underline{70.23} &  \underline{70.01} & 77.64 &  \underline{41.11} &  \textbf{46.89} &  \underline{25.26} & \underline{26.46} & 30.31 & 12.54 &  \underline{92.78} & 87.69 & \underline{86.12} \\
\multirow{-13}{*}{0\%Missing} & \textbf{CoCo-IMC(ours)} & \textbf{77.33} & \textbf{73.63} & \textbf{89.62} & \textbf{43.61} & \underline{46.52} & \textbf{26.64} & \textbf{27.84}  & \textbf{32.82} & \textbf{13.53} & \textbf{95.19} & 88.82 & \textbf{89.57} \\ \bottomrule
\end{tabular}}
\end{table}

We test all methods on a dataset with 50$\%$ missing rate and a complete dataset. As shown in Table. \ref{table4.2}, CoCo-IMC was clearly better than these state-of-the-art baselines on all four datasets. CoCo-IMC performance is better than methods within the complete data that do not participate in training, such as UEAD, DCCA, PIC and COMPLETER, which indicate that CoCo-IMC retains more complementary features. DAIMC, PVC, EERIMVC and PIC pursue consistency learning and data recovery. DCCAE, PVC, BMVC and PIC only contain consistency learning. DAIMC only focuses on complementary. CoCo-IMC not only considers complementarity and consistency but also achieves a balance between them. We can conclude that in multi-view clustering, balancing complementarity and consistency can improve the clustering performance.

\subsection{Parametric Analysis and Ablation Studies}
\label{sec4.5}
\begin{figure}[h!]
\centering
\subfigure[]{
\includegraphics[width=2.8cm]{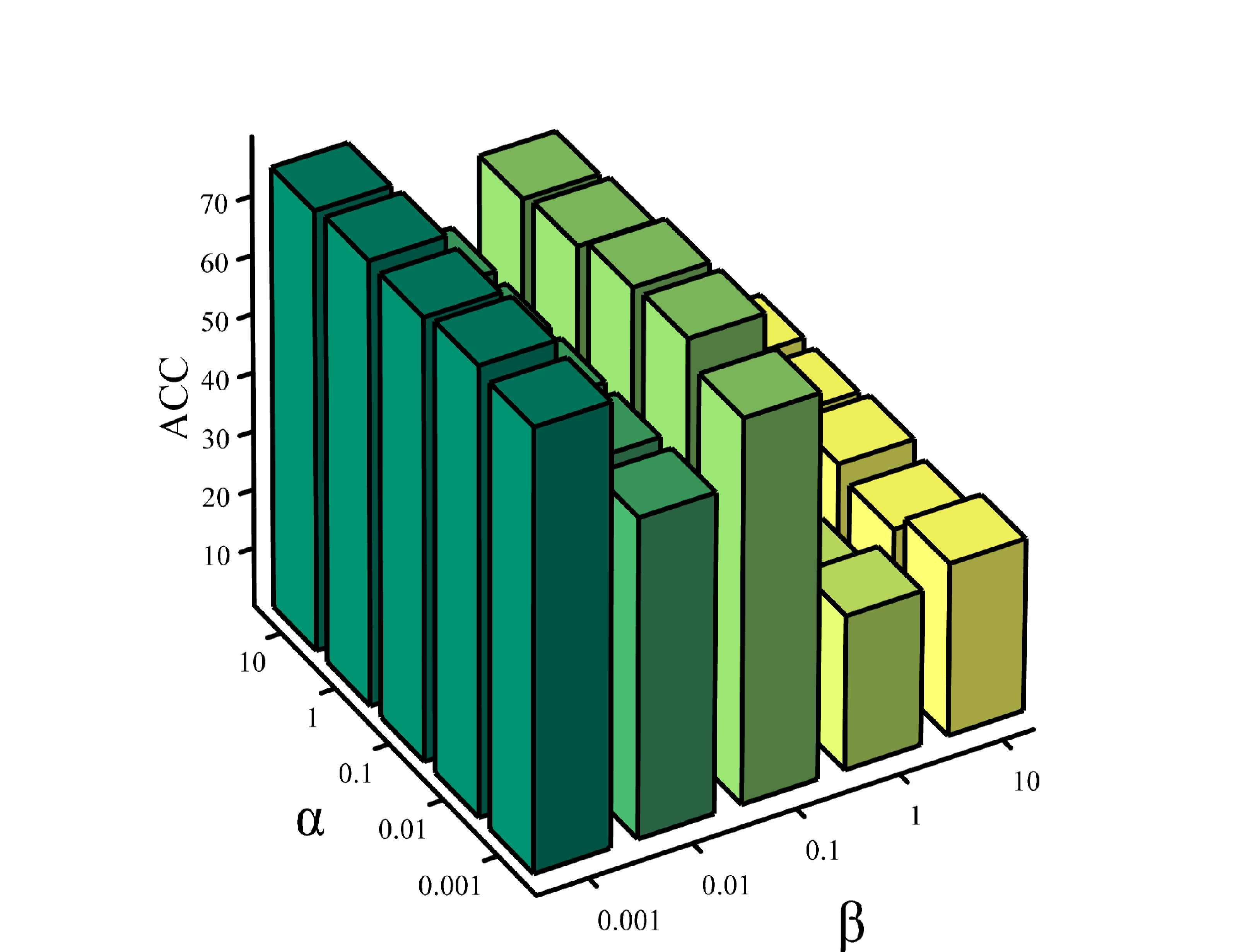}
}\subfigure[]{
\includegraphics[width=2.8cm]{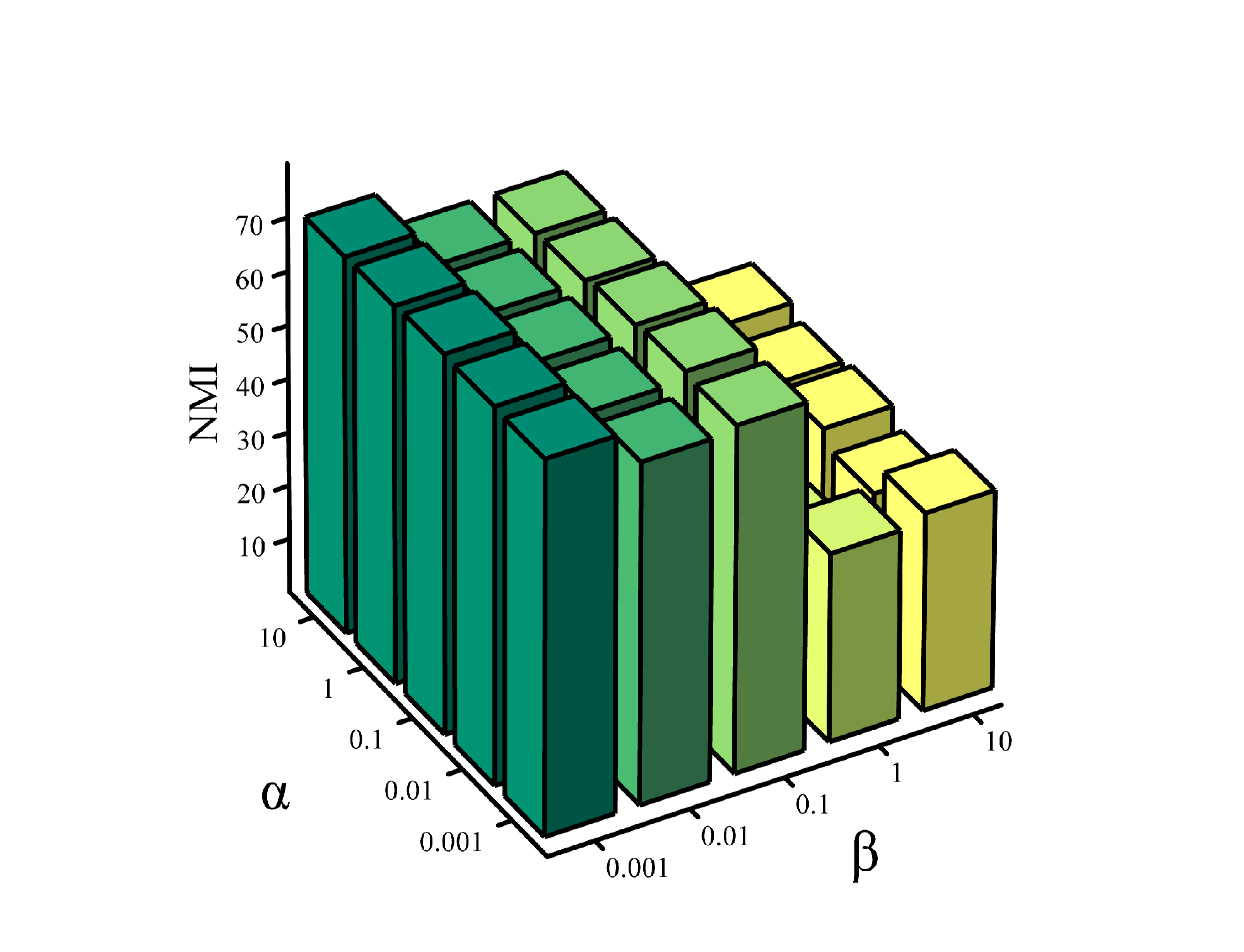}
}
\subfigure[]{
\includegraphics[width=2.8cm]{pdf/Fig7b.png}
}\subfigure[]{
\includegraphics[width=2.8cm]{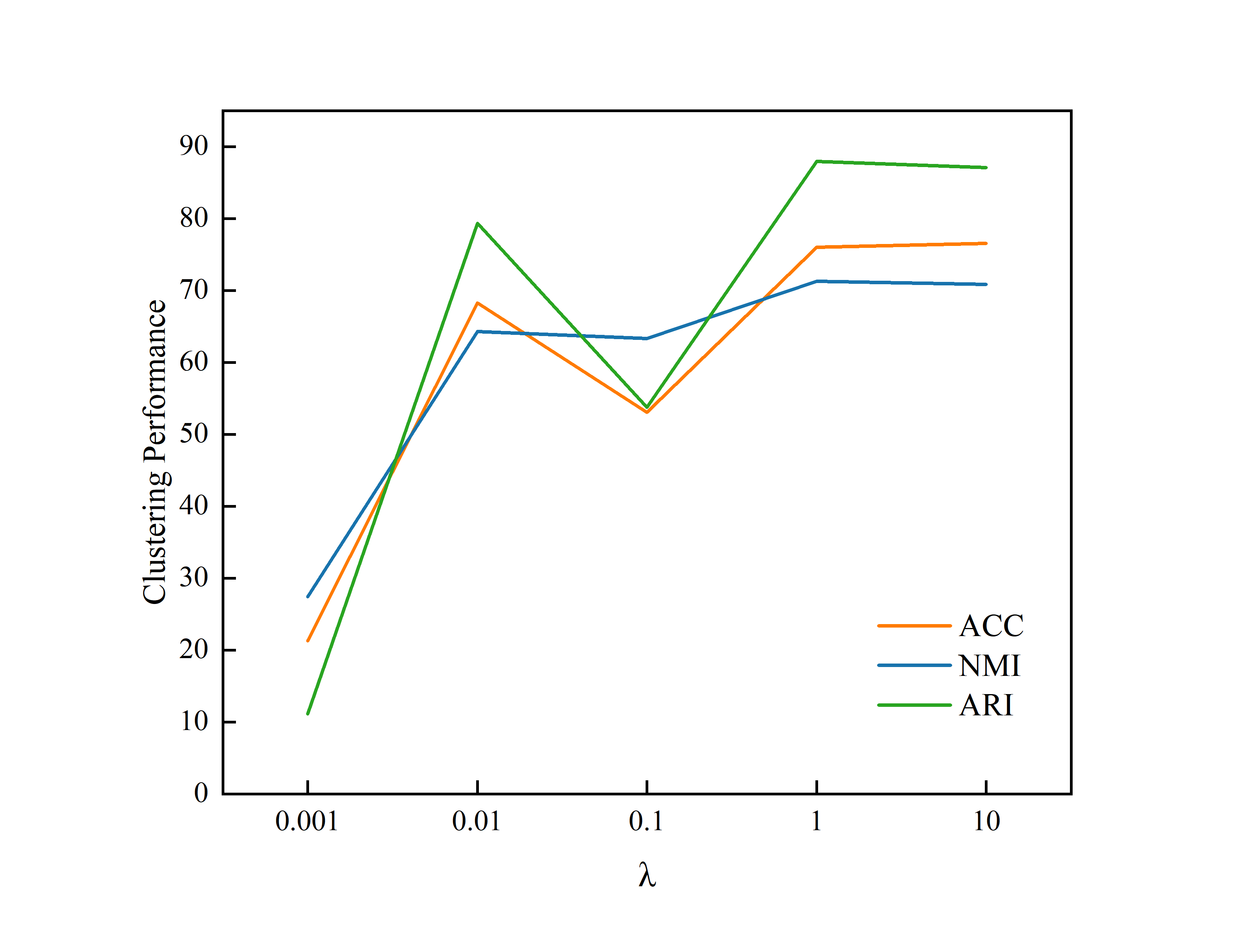}
}
\caption{Parameter sensitivity analysis on Caltech101-20.}
\label{fig7}
\end{figure}

We conduct experiments to assess CoCo-IMC's trade-off hyper-parameters on Caltech101-20. As shown in Fig. \ref{fig7}(a, b, c), with the missing rate fixed at 0.5, we evaluate $\alpha$, $\beta$ and $\lambda$, and we vary their values in the range of 0.001, 0.01, 0.1, 1 and 10. The results are best when the values of $\alpha$ and $\beta$ are set to 0.01, 0.001 by fixing $\lambda$ at 1. The results show that complementary grows gradually and the clustering performance (ACC, NMI and ARI) first improves and then decreases. The reason is that increased complementarity brings additional features, which may improve the clustering performance. However, too much complementary will negative consistency learning. As shown in Fig. \ref{fig7} (d), we fixed $\alpha$ and $\beta$ to 0.01 and 0.001, respectively, and studied $\lambda$. We discern a notable influence of $\lambda$ on the model's performance. Conversely, the impact of $\alpha$ and $\beta$ is relatively minor.
\vspace{-0.8cm}
\vspace{-0.8cm}

After that, we performed ablation studies to demonstrate the importance of each component of CoCo-IMC. Specifically, the Caltech101-20 dataset with a missing rate of 0.5 is selected for experiments, and 11 experiments are designed to separate the effects of consistency learning and complementary learning. Table \ref{table4.3} shows the results of four modules, which prove all losses play an key role in our method. It should be point out that optimizing $\mathcal{L}_{cnl}$ may result in trivial solutions. To address this problem, we must optimize it during view reconstruction process.

\subsection{Performance with Different Missing Rates and Different Momentum}
\label{4.3}
\begin{figure*}[htbp]
    \centering
    \begin{minipage}[t]{0.3\linewidth}
    \centering
    \includegraphics[width=1\linewidth]{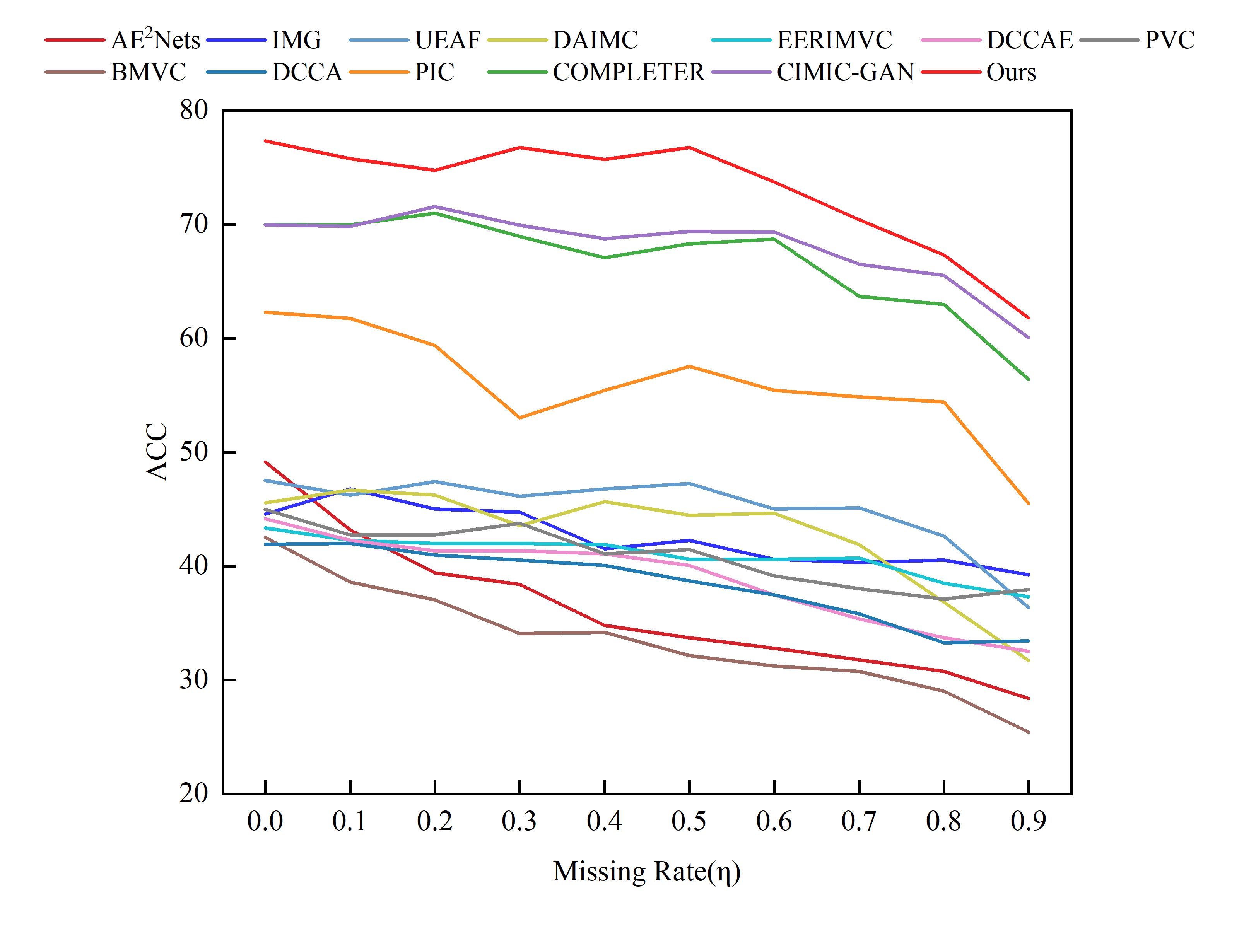}
    \caption{\small{Missing rates analysis.}}
    \label{fig5}
    \end{minipage}
    \quad
     \begin{minipage}[t]{0.3\linewidth}
    \centering
    \includegraphics[width=1\linewidth]{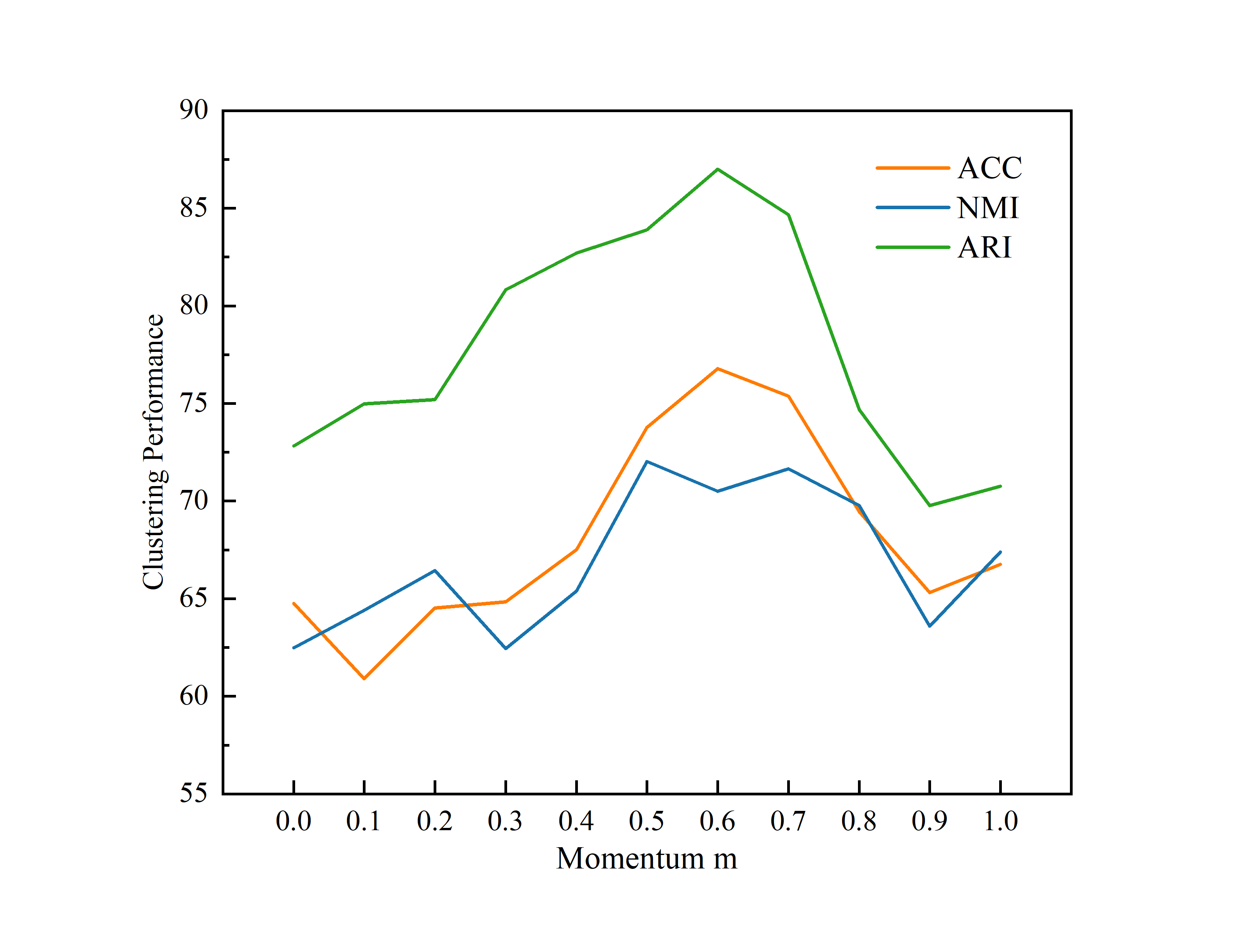}
    \caption{\small{Momentum analysis.}}
    \label{fig6}
    \end{minipage}
    \quad
    \begin{minipage}[t]{0.3\linewidth}
    \centering
    \includegraphics[width=1\linewidth]{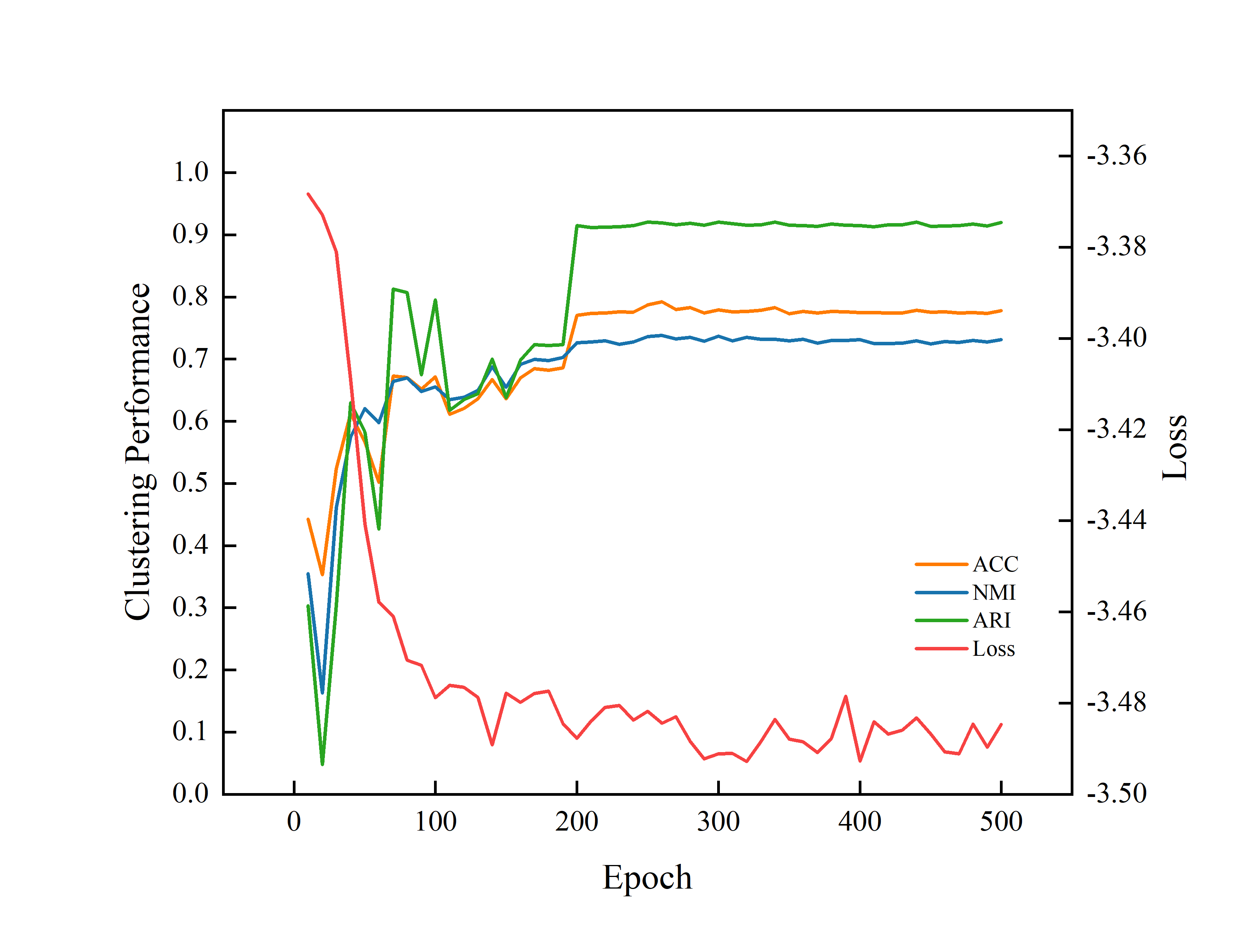}
    \caption{\small{Convergence analysis}}
    \label{fig9}
    \end{minipage}
\end{figure*}

In this section, we carry out experiments with different methods at different misssing rates $\eta$ to validate the  efficacy of CoCo-IMC. On the Caltech101-20 dataset, we cut the missing rate $\eta$ into 10 examples from 0 to 0.9 with an interval of 0.1.  As shown in Fig. \ref{fig5}, we could observe that. (i) AE$^2$Nets, DCCAE, BMVC and DCCA all rely heavily on view pairwise relationships, but when the missing rate is 1, this renders the above methods ineffective. Similarly, COMPLETER relies heavily on paired samples of complete data. When the missing rate is high, the model collapses, i.e., the accuracy is only 26.33$\%$ when all views are missing. (ii) EERIMVC and PIC, fill in the missing views by exploiting the similarity of the remaining views, which staill valid when the missing rate is 1. CIMIC-GAN performs well even in the absence of paired samples by directly inferring missing data. Therefore, we also achieve near-optimal performance by inferring missing data. In all missing rate settings, CoCo-IMC performed significantly better than the baseline for all tests. When the missing rate is 0.9, CoCo-IMC achieves an accuracy of 61.8$\%$, which is 1.72$\%$ higher than the sub-optimal methods. This experiment also demonstrates the strong robustness and generalization ability of the model.
\vspace{-0.6cm}
\begin{table}
\renewcommand\arraystretch{0.8}
\centering
\footnotesize
\caption{Ablation study on Caltech101‐20 with a missing rate of 0.5. In the table, "\Checkmark" denotes CoCo-IMC with the component.}
\label{table4.3}
\setlength{\tabcolsep}{4mm}{
\begin{tabular}{cccccccc}
\hline
\multicolumn{1}{l}{} & \multirow{2}{*}{$\mathcal{L}_{rec}$} & \multicolumn{1}{l}{\multirow{2}{*}{$\mathcal{L}_{cml}$}} & \multicolumn{2}{c}{$\mathcal{L}_{cnl}$} & \multirow{2}{*}{ACC} & \multirow{2}{*}{NMI} & \multirow{2}{*}{ARI} \\
 &  & \multicolumn{1}{l}{} & $\mathcal{L}_{pre}$ & $\mathcal{L}_{ccl}$ &  &  &  \\ \hline
(1) & \Checkmark &  &  &  & 31.22 & 45.20 & 21.84 \\
(2) &  & \Checkmark &  &  & 21.03 & 12.73 & 3.22 \\
(3) &  &  & \Checkmark &  & 36.75 & 18.10 & 10.15 \\
(4) &  &  &  & \Checkmark & 44.14 & 58.57 & 41.03 \\
(5) & \Checkmark & \Checkmark &  &  & 30.72 & 41.97 & 20.78 \\
(6) & \Checkmark &  & \Checkmark &  & 44.99 & 61.52 & 36.37 \\
(7) & \Checkmark &  &  & \Checkmark & 58.78 & 61.12 & 66.87 \\
(8) & \Checkmark & \Checkmark & \Checkmark &  & 44.70 & 60.86 & 33.36 \\
(9) & \Checkmark & \Checkmark &  & \Checkmark & 61.67 & 62.88 & 66.20 \\
(10) & \Checkmark &  & \Checkmark & \Checkmark & 69.21 & 70.24 & 82.31 \\
(11) & \Checkmark & \Checkmark & \Checkmark & \Checkmark & 76.78 & 70.51 & 87.01 \\ \hline
\end{tabular}
}
\end{table}
\vspace{-0.4cm}
After that, we implement experiments to validate the efficacy of delayted activate by setting a different momentum $m$  on the Caltech101-20 dataset. As shown in Fig. \ref{fig6}, We could observe that if the online network parameters $\theta$ updated in the current step are directly copied to the target network will eventaually be completely disorderd, causing the incapacity to train to oscillate. When $m$ is small, the target network parameters are updated too quickly and the complementary will be ignored, while the contrary will affect the consistentency. Therefore, too fast or too slow parameter changes are not beneficial to clustering. So, introducing momentum $m$ can help us find the most suitable balance.  
\vspace{-0.4cm}
\vspace{-0.8cm}

\subsection{Convergence Analysis and Visualisation}
\label{sec4.7}
In this section, we investigate the convergence of CoCo-IMC by reporting the loss value and the clustering performance with increasing epochs. As shown in Fig. \ref{fig9}, one could observe that the loss remarkably decreases in the first 200 epochs, ACC, NMI and ARI continuously increase and tend to be smooth and consistent.

\begin{figure}[h!]
\centering
\subfigure[Caltech101-20]{
\includegraphics[width=2.5cm]{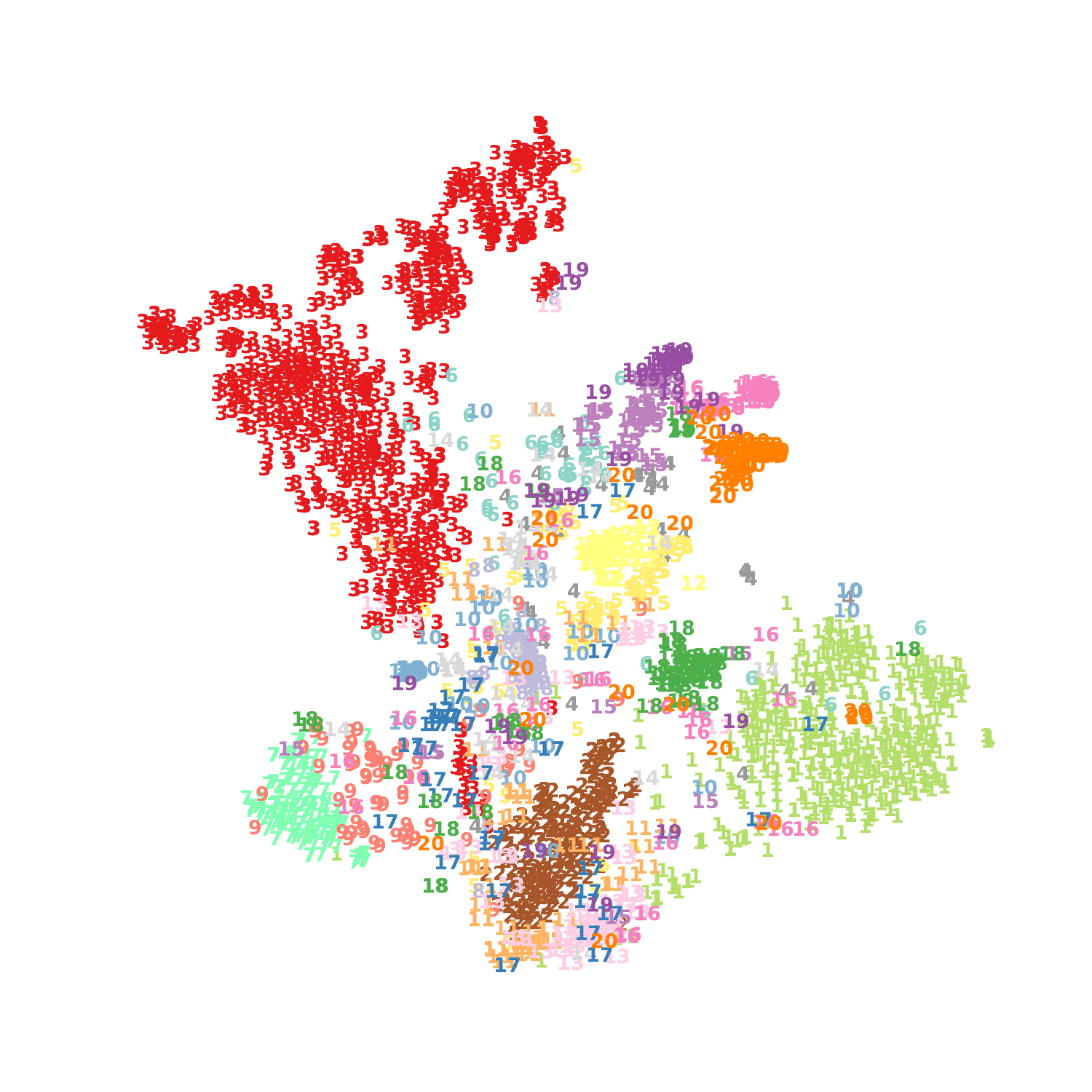}
\includegraphics[width=2.5cm]{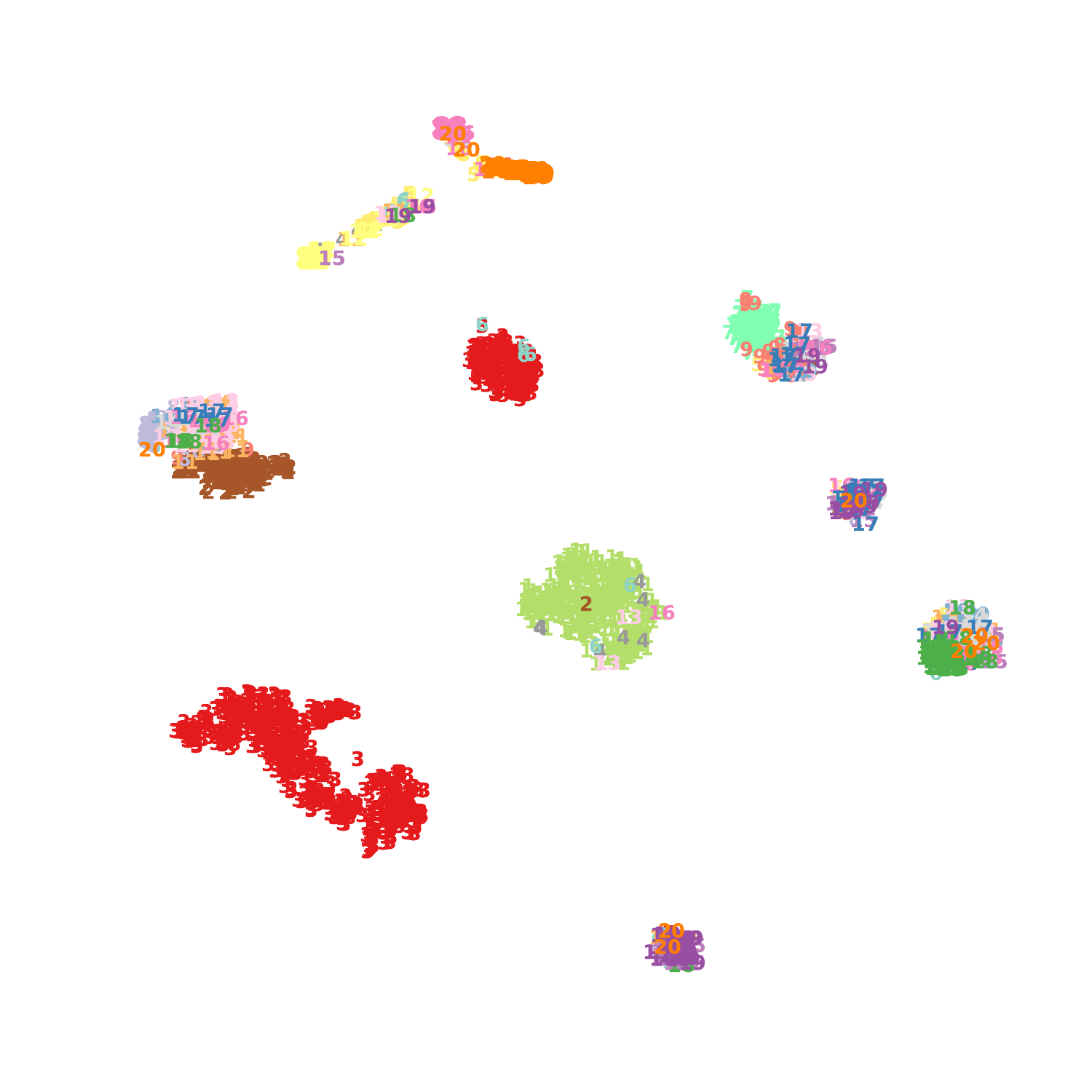}
}\subfigure[Scene-15]{
\includegraphics[width=2.5cm]{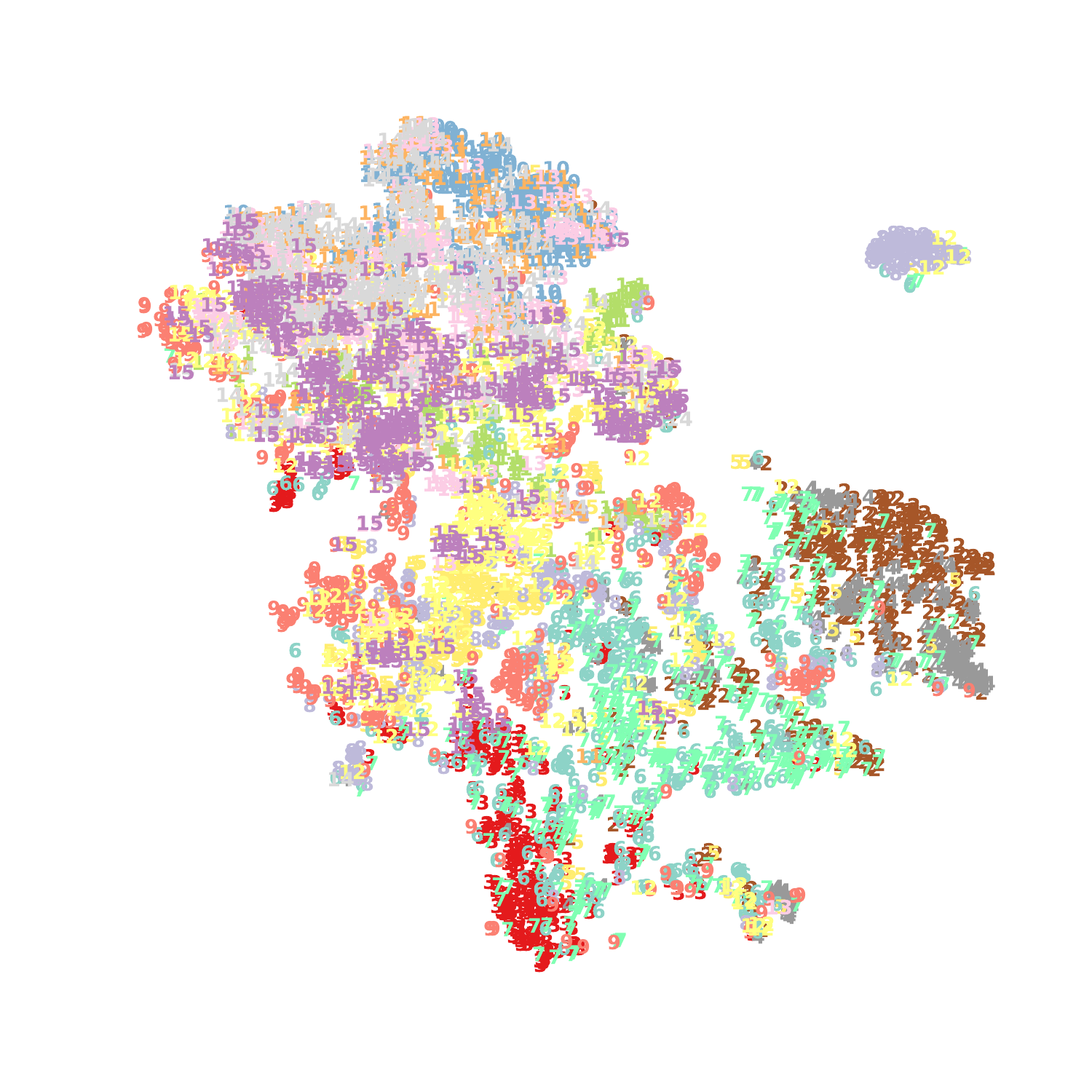}
\includegraphics[width=2.5cm]{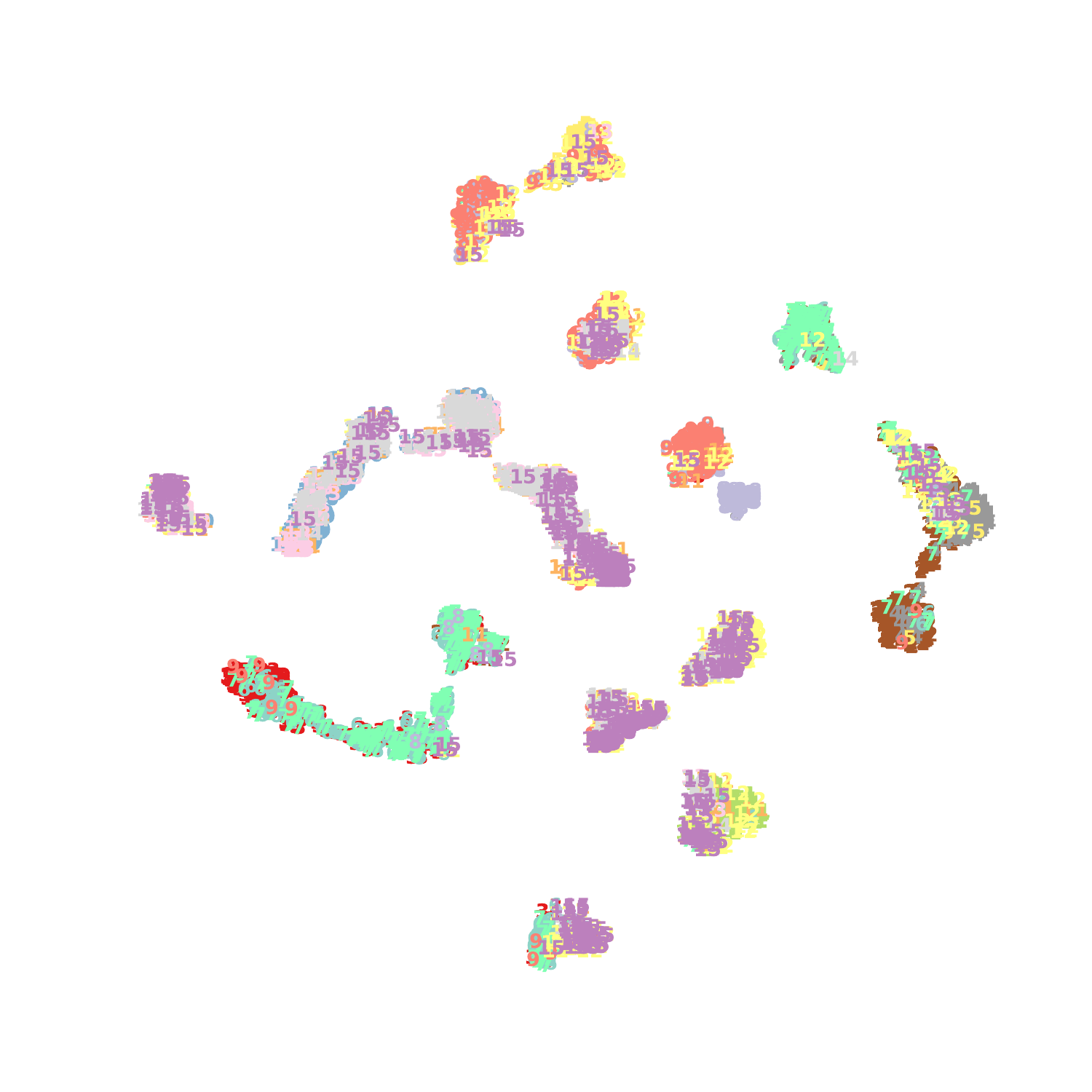}
}
\subfigure[LandUse-21]{
\includegraphics[width=2.5cm]{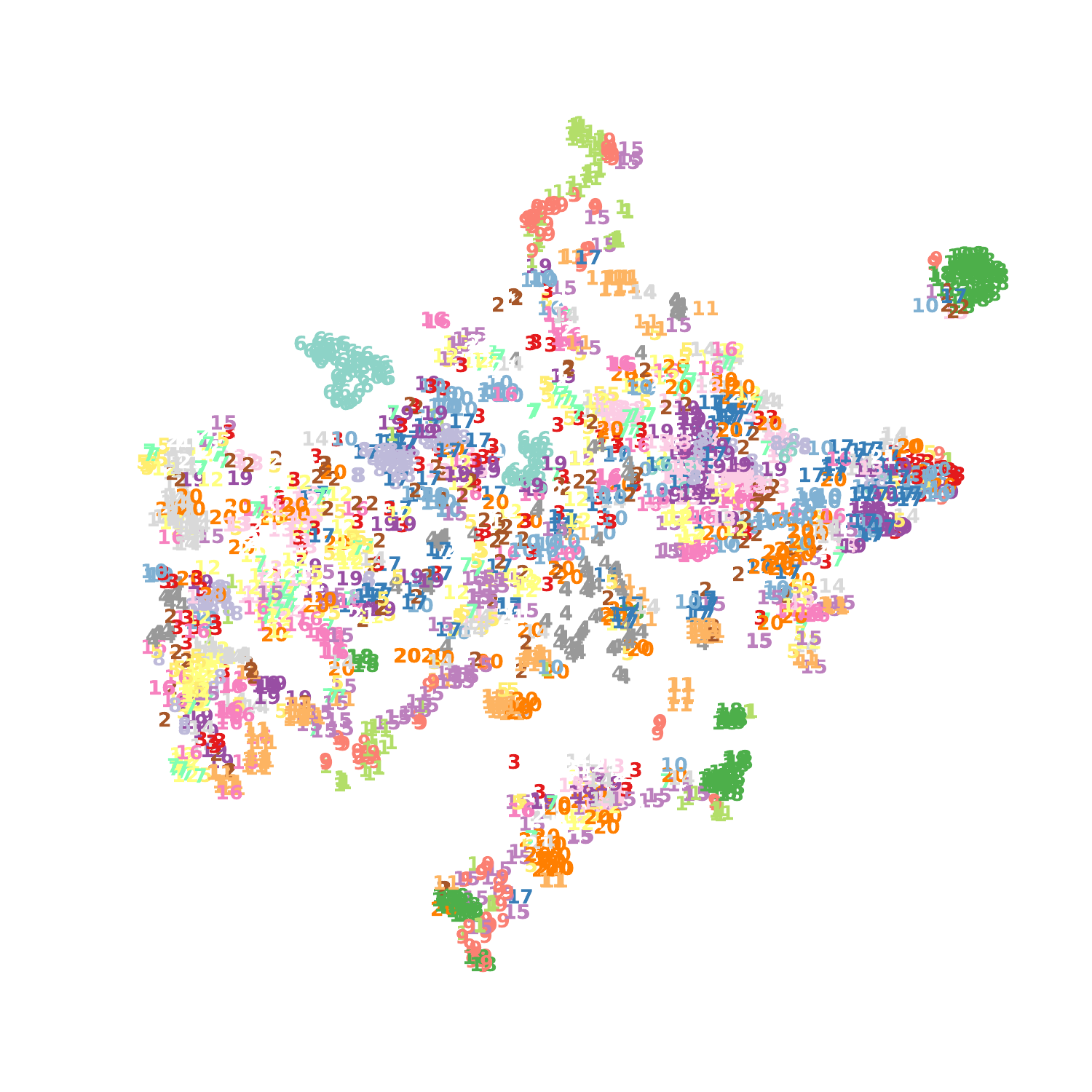}
\includegraphics[width=2.5cm]{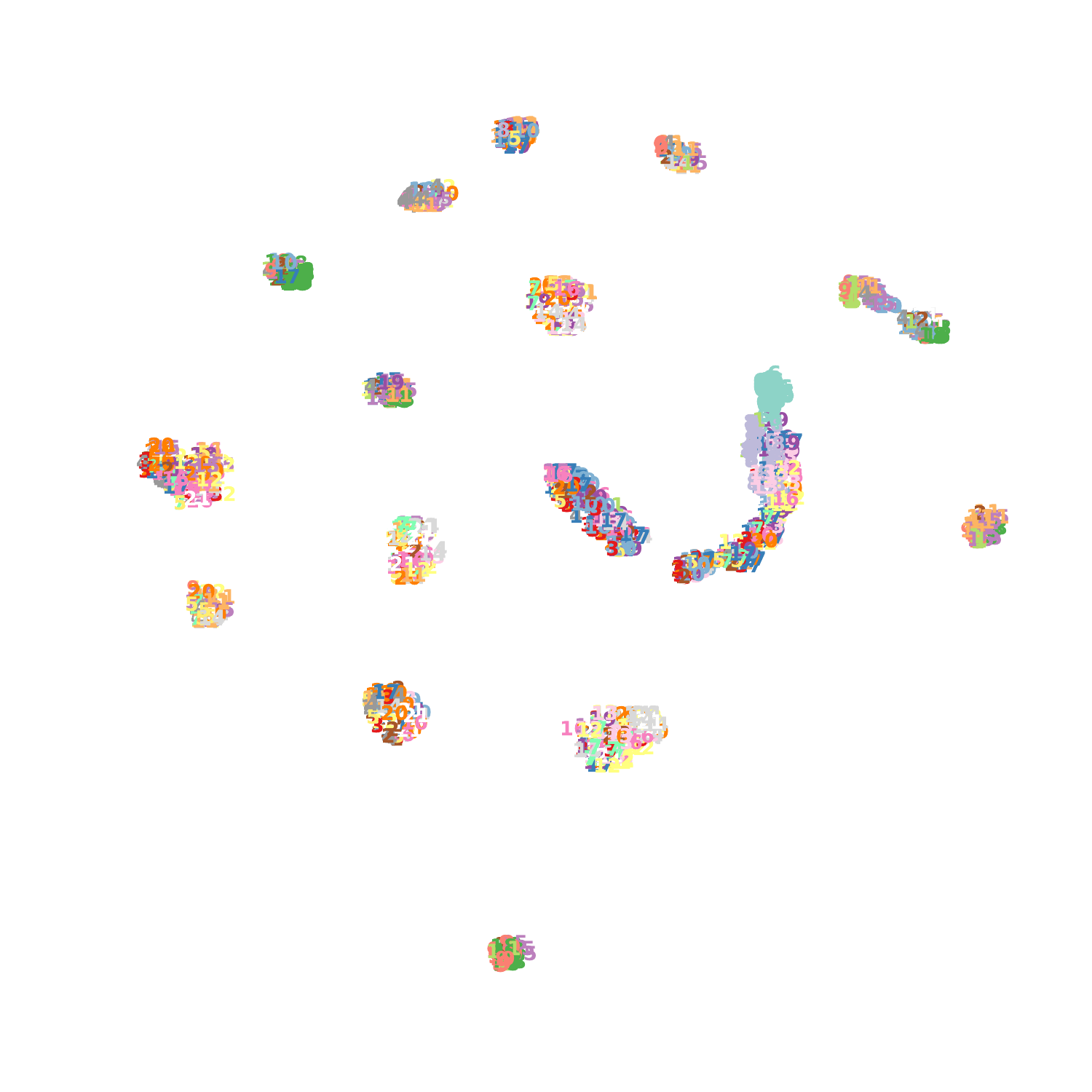}
}\subfigure[Noisy MNIST]{
\includegraphics[width=2.5cm]{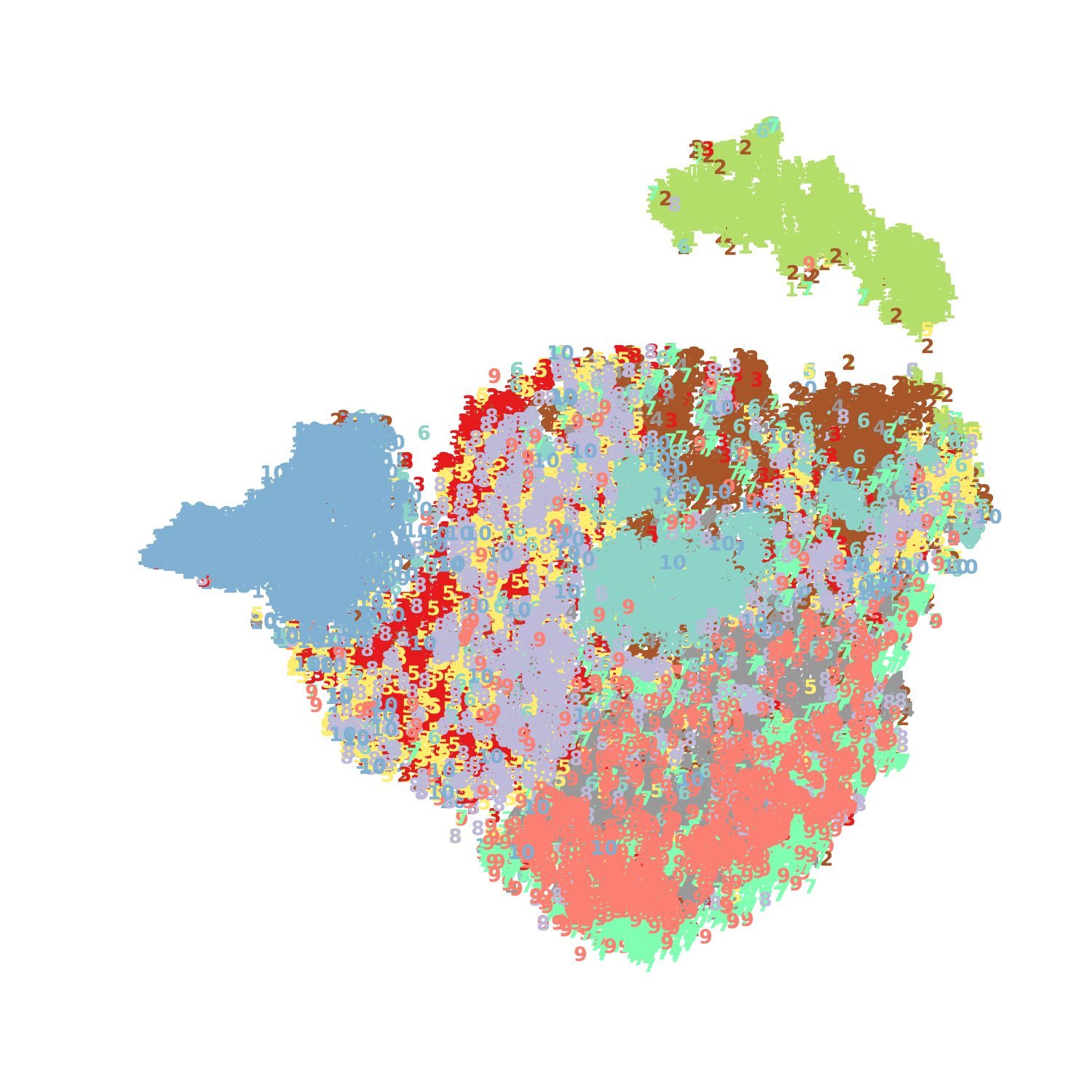}
\includegraphics[width=2.5cm]{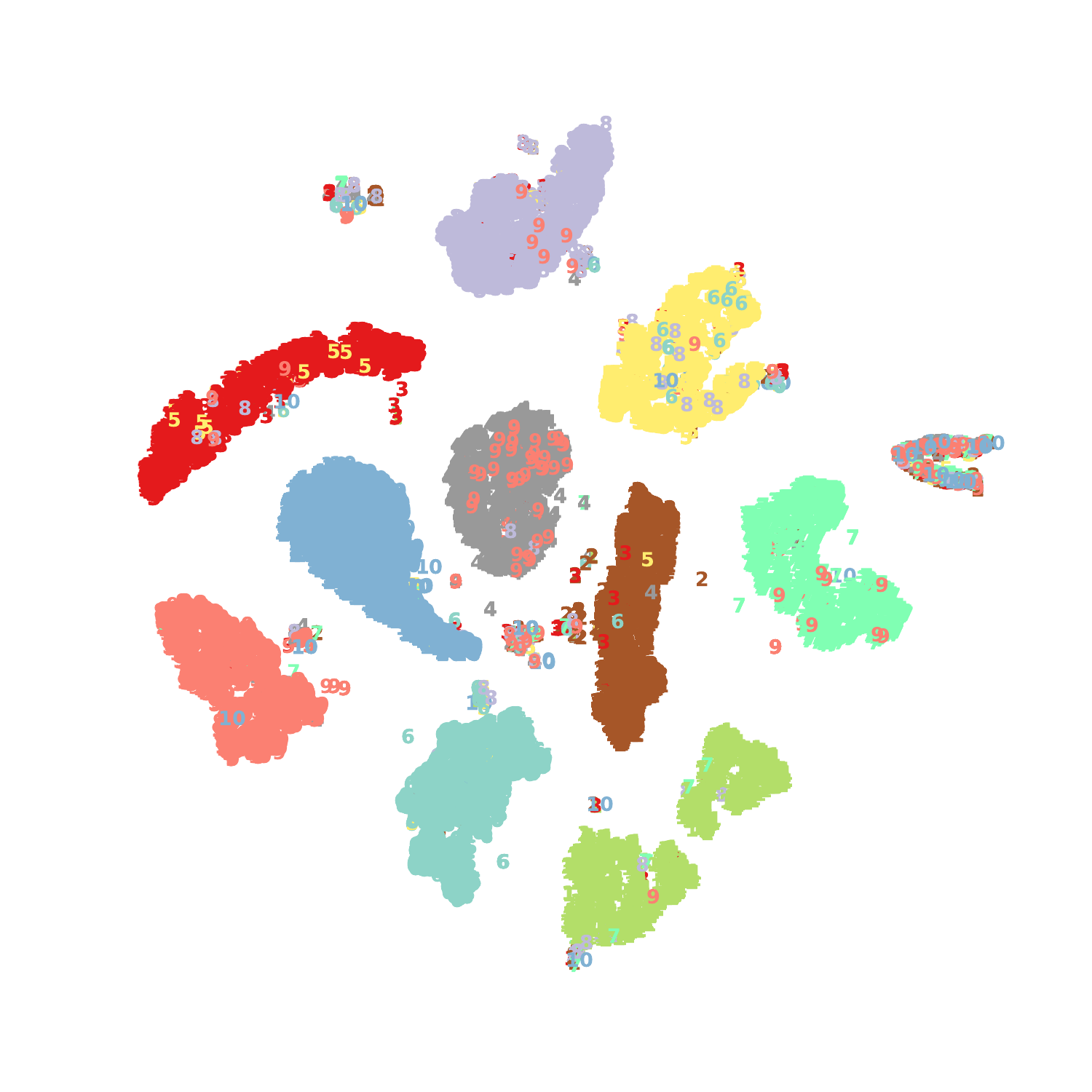}
}
\caption{\small{Visualization on four datasets via t-SNE \cite{106} with the initial and final situation.
}}
\label{fig10}
\end{figure}

To demonstrate the superiority of the CoCo-IMC intuitively, we conduct the t-SNE visualizations of the representations learned on the four datasets with a missing rate of 0.5. As shown in Fig. \ref{fig10}, we conclude that, compared with the raw features, CoCo-IMC could achieve more compactness and independence from clusters.
\leavevmode\newline\vspace{-1cm}

\vspace{-1cm}
\section{CONCLUSION}
\label{sec5}
This paper proposes CoCo-IMC to provide a unifying framework for handling the challenging problem of incomplete multi-view clustering, which makes it hard to balance complementarity and consistency. In short, complementarity and consistency of multi-view are positive and negative aspects to be considered together based on application or a priori knowledge, which are not two separate issues. Such a unified framework would provide novel insight to the community on balancing complementarity and consistency. In the future, we hope the framework will handle more practical scenarios. In industrial-grade scenarios, incomplete information is always unavoidable. It should be especially emphasized that the complementary information is very important, especially in domains such as unmanned aerial vehicle and autonomous vehicleserless.

\subsubsection{Acknowledgements.} This work is supported by the National Natural Science Foundation of China (No. 62062045, No. 61962055), Young Talents of Science and Technology in Universities of Inner Mongolia Autonomous Region (NJYT23104, NJYT24061) and the Natural Science Foundation of Inner Mongolia Autonomous Region (JY20220249, JY20240061).
\bibliographystyle{IEEEtran}
\bibliography{References}
%




\end{document}